\documentclass[11pt,letterpaper]{article}

\usepackage[margin=1in]{geometry}
\usepackage{amsmath,amssymb,amsthm}
\usepackage{graphicx}
\usepackage{booktabs}
\usepackage{array}
\usepackage{listings}
\usepackage{xcolor}
\usepackage{enumitem}
\usepackage{caption}
\usepackage{tikz}
\usetikzlibrary{shapes.geometric,arrows.meta,positioning}
\usepackage[hidelinks]{hyperref}
\hypersetup{
  pdftitle={Self-Emergence Agent Architecture: Behavior-Inertia HMM, Reflexive Metacognition, and Social-Contrastive Self-Modeling},
  pdfauthor={Xiaoyang Liu},
  pdfsubject={LLM agent architecture and self-emergence},
  pdfkeywords={LLM agents, hidden Markov model, behavioral inertia, metacognition, multi-agent interaction, self-emergence}
}

\usepackage{etoolbox}
\usepackage{placeins}
\pretocmd{\section}{\FloatBarrier}{}{}

\definecolor{codegray}{gray}{0.97}
\definecolor{codekw}{RGB}{0,70,140}
\definecolor{codecmt}{RGB}{90,110,90}
\newtheorem{hypothesis}{Hypothesis}
\newtheorem{definition}{Definition}

\title{\textbf{Self-Emergence Agent Architecture:\\
Behavior-Inertia HMM, Reflexive Metacognition,\\
and Social-Contrastive Self-Modeling}}

\author{Xiaoyang Liu\\
Independent Researcher\\
\texttt{liuxiaoyang\_1@outlook.com}}

\date{\today}

\begin{document}
\maketitle

\begin{abstract}
Large language model (LLM) agents exhibit strong language-generation and
problem-solving capabilities, yet they suffer from three structural
limitations: \emph{personality drift}, \emph{non-evolutionary reflection}, and
the \emph{absence of a self--other boundary}. Existing generative-agent
simulations rely on static memory and fixed prompt constraints; they neither
maintain continuous individual behavioral inertia nor realize endogenous
self-evolution. We propose the \textbf{Self-Emergence Agent Architecture
(SEAA)}, which integrates three components: (i) a Hidden Markov Model (HMM)
that encodes long-term behavioral and cognitive inertia as an editable
state-transition matrix; (ii) a Reflexion-style verbal metacognition loop whose
output \emph{updates the HMM parameters themselves}, rather than merely being
stored as text; and (iii) a multi-agent social environment in which initially
identical agents continuously compare their behavior with that of others. The
three components form a closed loop: social action $\rightarrow$ feedback
$\rightarrow$ self-reflection $\rightarrow$ inertia update $\rightarrow$
differentiated action. We state three falsifiable hypotheses---individual
differentiation, reflection-driven evolution, and social-contrastive boundary
formation---and provide a complete, reproducible experimental protocol with
operational metrics. A language-model-free mechanism prototype shows that the
closed loop spontaneously breaks symmetry: initially identical agents
consolidate distinct, stable personalities whereas matched controls do not.
Experiments with a real hosted LLM then surface these internal differences as
distinct first-person self-narratives, and a five-agent deliberation
spontaneously develops social structure---a consensus hub and a unanimously
rejected outlier---that never appears in the undifferentiated control.
Following an epistemologically agnostic stance inspired by Zhuangzi, SEAA
studies only \emph{observable behavioral emergence} and makes no claim about
subjective qualia. This work contributes a unified theoretical framework, a
concrete agent architecture with pseudocode, preliminary mechanistic evidence,
and a microscope-style experimental sandbox for studying the emergence of an
artificial self.
\end{abstract}


% ============================================================
\section{Introduction}\label{sec:intro}
% ============================================================

Transformer-based large language models have achieved breakthroughs in natural
language processing, reasoning, and interactive simulation. Building on them,
generative agents such as the Stanford ``smallville'' simulacra
\cite{park2023generative} and self-reflective agents such as Reflexion
\cite{shinn2023reflexion} can now emulate human dialogue, sequential
decision-making, and elementary social behavior.

When the goal shifts from solving a single task to emulating an
\emph{individualized, developing self}, however, current LLM agents exhibit
three fundamental bottlenecks.

\paragraph{(1) Lack of stable individual inertia.}
A base LLM reflects the average distribution of a global public corpus rather
than the unique life trajectory of an individual. Without a persistent
internal state, an agent's personality and behavioral logic drift with the
prompt, and it cannot exhibit the cross-time consistency that characterizes a
human individual.

\paragraph{(2) Reflection that does not transform the self.}
Mainstream reflection frameworks store a natural-language summary in context
memory to improve the \emph{next task}, but they never modify the agent's
underlying behavioral disposition. The agent can ``summarize a lesson'' yet
cannot ``change who it is,'' lacking the growth loop in which human reflection
reshapes long-term tendencies.

\paragraph{(3) No endogenous self--other boundary.}
Existing multi-agent simulations realize behavioral exchange but do not
implement a mechanism through which an agent distinguishes itself from others.
Human self-concept is not given at birth; it emerges through prolonged social
comparison and the recognition of behavioral difference---a process absent from
current artificial societies.

\subsection{Motivation and philosophical stance}
We posit two origins of human selfhood: \emph{individual behavioral and
cognitive inertia} accumulated over a personal history, and a
\emph{self--other boundary} generated by contrasting one's own conduct with
that of others during social interaction.

We adopt an epistemologically cautious stance articulated in the classical
Chinese text \emph{Zhuangzi}: ``You are not a fish; how do you know the joy of
fish?'' Subjective experience cannot be directly inspected from the outside.
What an external observer \emph{can} measure is continuous, differentiated, and
evolving behavior. Accordingly, SEAA treats \emph{observable behavioral
emergence} as its sole object of study---analogous to watching paramecia under a
microscope---and deliberately avoids asserting the presence or absence of
subjective qualia.

\subsection{Contributions}
This paper makes three contributions.
\begin{enumerate}[leftmargin=2em,itemsep=2pt]
  \item \textbf{Mathematical carrier of individuality.} We encode persistent
  behavioral/cognitive inertia as an HMM transition matrix, turning abstract
  personality continuity into a computable and editable object.
  \item \textbf{Evolutionary reflection loop.} Verbal metacognition is parsed
  into parameter-level updates that modify the HMM itself, realizing endogenous
  self-change rather than text-only memorization.
  \item \textbf{Social-contrastive self-emergence.} Grounded in symbolic
  interactionism and the looking-glass self, initially homogeneous agents
  spontaneously differentiate in interaction and form a self--other boundary.
\end{enumerate}

\subsection{Organization}
Section~\ref{sec:related} reviews related work and positions SEAA against
Reflexion and Generative Agents. Section~\ref{sec:hypotheses} states our
hypotheses. Section~\ref{sec:architecture} details the architecture and
pseudocode. Section~\ref{sec:experiments} gives the protocol together with the
mechanism prototype and real-LLM results. Section~\ref{sec:discussion}
discusses scope and philosophy, and Section~\ref{sec:conclusion} concludes.
Appendices~A--D collect the exact prompts, verbatim self-interview and
deliberation transcripts, and full reproducibility details.

% ============================================================
\section{Related Work}\label{sec:related}
% ============================================================

\subsection{Self-reflective and self-evolving LLM agents}
Reflexion \cite{shinn2023reflexion} introduces verbal reinforcement learning:
after a task, the agent records its trajectory, evaluates the outcome, writes a
natural-language reflection, and reuses it---without gradient updates. ReAct
\cite{yao2023react} interleaves reasoning traces with actions. Recent surveys
of \emph{self-evolving agents}
\cite{xiang2026survey,gao2025survey} organize a rapidly growing literature in
which agents improve their prompts, memories, tools, or parameters across
interactions. Almost all of this work optimizes \emph{task performance};
the object that evolves is a skill, a memory store, or a policy. What does not
evolve is an enduring \emph{behavioral disposition}: these agents can
``summarize a lesson'' yet cannot ``change who they are.'' SEAA is orthogonal
to this literature---the variable being updated is a personality carrier, not
a competence carrier.

\subsection{Generative multi-agent simulation and emergent individuality}
Generative Agents \cite{park2023generative} demonstrate believable daily social
behavior through memory retrieval and prompt-based planning; Humanoid Agents
\cite{wang2023humanoid} add System-1-style needs and emotions on top of a
similar architecture. In these systems each agent's persona is hand-authored
and fixed. A complementary line asks whether individuality can emerge from an
\emph{undifferentiated} start: Takata, Masumori, and Ikegami
\cite{takata2024individuality} report that initially identical LLM agents
differentiate behavior, emotion, and personality through group communication;
their El Farol study \cite{takata2025elfarol} further shows spontaneous role
differentiation and bounded-rational behavior in a classic social dilemma.
Lai et al.\ \cite{lai2024collectives} show that freely interacting AI
collectives develop divergent emergent subjectivities, and Zhang et al.\
\cite{zhang2026stances} report that agents in generative societies form
endogenous stances that can override their preset identities and re-organize
community boundaries. These studies establish the \emph{phenomenon} of
emergent individuality observationally, but their agents keep a fixed internal
logic: there is no editable inertia parameter and no reflection-driven
self-evolution loop, so the mechanism \emph{behind} the differentiation
remains implicit, and long-term personality growth cannot be modeled,
controlled, or ablated.

\subsection{Markov models of behavior, affect, and personality}
Hidden Markov Models \cite{rabiner1989tutorial} are widely used for sequence
prediction, affect recognition, and trait simulation, and adaptive/online
variants can update transition probabilities from numerical signals. The
closest prior combination with LLMs is MECoT \cite{wei2025mecot}, which drives
emotionally consistent role-playing with a \emph{personality-weighted} Markov
chain over an emotion circumplex, coupled to LLM reasoning. The crucial difference is what flows \emph{into}
the matrix: in MECoT and related systems the transition structure is estimated
or authored \emph{before} deployment and stays fixed at run time, whereas in
SEAA the matrix is \emph{edited at run time} by the agent's own natural-language
metacognition. To our knowledge, no prior work drives HMM inertia updates from
LLM verbal self-reflection for the purpose of self-emergence.

\subsection{Evolving-persona and lifelong agents}
A parallel engineering literature builds persistent, evolving personas:
Generative Life Agents \cite{gla2025} maintain a Perceive--Retrieve--Reflect--Evolve--Plan--Act
loop in which a meta-cognitive process explicitly rewrites the agent's traits,
goals, and interests; AutoPersonas \cite{autopersonas2026} attacks
persona collapse with a multi-timescale state-revision engine. Diagnostic
studies quantify how LLM personas shift under major life events
\cite{wang2026personasgrow}, finding that current agents reproduce the
\emph{mean} of human personality dynamics but not its shape.
These systems show that reflection-driven trait edits are feasible and useful,
but the edited object is a \emph{scalar trait vector or identity document},
not a stochastic model of behavioral inertia; they are single-agent by design
and contain no social mechanism through which a self--other boundary could
form endogenously.

\subsection{Theories of consciousness and machine self-models}
The functionalist view holds that consciousness is an informational architecture
independent of substrate, whereas biological accounts tie subjective experience
to living tissue. Computational proposals such as Global Workspace Theory and
Predictive Processing \cite{friston2010free,dehaene2017consciousness} target
specific mechanisms, and Butlin et al.\ \cite{butlin2023consciousness} derive
indicator properties for consciousness in current AI systems; Metzinger's
self-model theory \cite{metzinger2003being} grounds subjectivity in a system's
transparent model of itself, and Seth \cite{seth2021being} grounds selfhood in
embodied, predictive regulation. None of these frameworks integrates inertia,
self-reflection, and social contrast into one closed developmental loop---which
is the specific gap SEAA addresses at the behavioral level.

\subsection{Research gap}
Prior work treats inertia modeling, self-reflection, persona evolution, and
social interaction in isolation. Table~\ref{tab:comparison} contrasts SEAA
with representative lines; no prior public work realizes the full closed loop
\emph{social behavior $\rightarrow$ self-reflection $\rightarrow$ inertia
update $\rightarrow$ differentiation $\rightarrow$ self-emergence}.

\begin{table}[htbp]
\centering\footnotesize
\setlength{\tabcolsep}{6pt}
\caption{Positioning of SEAA relative to representative prior agent lines.
$\checkmark$ = supported, $\times$ = absent/fixed, -- = not applicable,
P = partial. ``Identical start'' means agents begin with no authored persona
and diverge endogenously; ``Refl.$\to$params'' means reflection structurally
updates an internal state rather than being stored only as text. Reflexion
keeps reflection as text only; Generative Agents use hand-authored personas;
MECoT weights a Markov chain by a \emph{fixed} personality; Generative Life
Agents edit scalar traits rather than a stochastic inertia model; Takata et
al.\ observe emergent individuality without an editable mechanism.}
\label{tab:comparison}
\begin{tabular}{lcccc}
\toprule
\textbf{Work} & \shortstack{\textbf{Editable}\\\textbf{inertia}} & \shortstack{\textbf{Refl.}$\to$\\\textbf{params}} & \shortstack{\textbf{Identical start,}\\\textbf{emergent persona}} & \shortstack{\textbf{Social-contrast.}\\\textbf{self-model}} \\
\midrule
ReAct \cite{yao2023react}                       & --          & --       & --       & -- \\
Reflexion \cite{shinn2023reflexion}            & --          & $\times$ & $\times$ & $\times$ \\
Generative Agents \cite{park2023generative}    & $\times$    & $\times$ & $\times$ & P \\
Adaptive HMM \cite{rabiner1989tutorial}        & \checkmark  & $\times$ & --       & $\times$ \\
MECoT \cite{wei2025mecot}                      & $\times$ (fixed) & $\times$ & $\times$ & $\times$ \\
Generative Life Agents \cite{gla2025}          & $\times$ (traits) & \checkmark & $\times$ & $\times$ \\
Takata et al.\ \cite{takata2024individuality}  & $\times$    & $\times$ & \checkmark & P \\
\textbf{SEAA (ours)}                           & \textbf{\checkmark (HMM)} & \textbf{\checkmark} & \textbf{\checkmark} & \textbf{\checkmark} \\
\bottomrule
\end{tabular}
\end{table}

Concretely, our control condition operationalizes the Reflexion regime: verbal
reflection is produced and stored at every step but is never allowed to edit
internal parameters, whereas the SEAA condition feeds the same reflection back
into the HMM. The two curves in every result figure are therefore a direct
mechanism-level comparison between text-only reflection (Reflexion-style) and
parameter-editing reflection (SEAA), not merely an enabled/disabled toggle.
Relative to Generative Agents, SEAA starts all agents from \emph{identical}
states with no hand-written personas and asks whether individuality emerges
endogenously, rather than scripting distinct personas up front.

% ============================================================
\section{Core Theoretical Hypotheses}\label{sec:hypotheses}
% ============================================================

We integrate cognitive psychology, symbolic interactionism
\cite{mead1934mind,cooley1902human}, and social-comparison theory
\cite{festinger1954theory} into three falsifiable hypotheses. These are
theoretical claims whose validity is tested by the protocol in
Section~\ref{sec:experiments}; they are not presented as established results.

\begin{definition}[Behavioral inertia]
Behavioral inertia is the conditional dependence of an agent's current
cognitive state on its prior states, represented by an HMM transition operator
$P_t$. A stable $P_t$ corresponds to a consistent disposition; its gradual
differentiation across agents constitutes individuality.
\end{definition}

\begin{hypothesis}[Individual differentiation]\label{hyp:diff}
When initially identical agents share a social environment, their HMM
transition matrices diverge over time as a function of distinct experiences,
and pairwise behavioral similarity decreases monotonically (up to a floor).
\end{hypothesis}

\begin{hypothesis}[Reflection-driven evolution]\label{hyp:refl}
Agents whose reflections update their HMM parameters adapt faster to
environmental change and converge to more stable personalities than matched
agents whose reflections are stored as text only.
\end{hypothesis}

\begin{hypothesis}[Social-contrastive boundary]\label{hyp:boundary}
Agents embedded in a multi-agent society progressively articulate explicit
distinctions between self and others in their self-descriptions; isolated
control agents do not.
\end{hypothesis}

\paragraph{Philosophical boundary.}
Consistent with the Zhuangzian position, SEAA does not test for qualia. It
operationalizes ``self-concept'' strictly through measurable behavior---matrix
divergence, consistency, and contrastive self-description---which is the only
level at which external claims are verifiable.

% ============================================================
\section{System Architecture}\label{sec:architecture}
% ============================================================

\subsection{Overview}
\begin{figure}[htbp]
\centering
\begin{tikzpicture}[
  font=\footnotesize,
  box/.style={rounded corners=2pt,draw=black!70,line width=0.5pt,
    align=center,minimum width=2.55cm,minimum height=0.95cm,fill=blue!6,inner sep=3pt},
  soc/.style={box,fill=purple!8},
  refl/.style={box,fill=green!9},
  core/.style={box,fill=orange!16,line width=0.9pt,draw=orange!75!black,font=\footnotesize\bfseries},
  flow/.style={-{Stealth[length=2.2mm]},line width=0.7pt,black!75},
]
\node[box]  (n1) at (90:2.5)   {1. Social\\Observation};
\node[box]  (n2) at (30:2.5)   {2. HMM Inertia\\State $z_t$};
\node[box]  (n3) at (-30:2.5)  {3. Action Gen.\\(memory, self-model)};
\node[soc]  (n4) at (-90:2.5)  {4. Multi-Agent\\Interaction \& Feedback};
\node[refl] (n5) at (-150:2.5) {5. Reflexive\\Metacognition $R_t$};
\node[core] (n6) at (150:2.5)  {6. Update $P_{t+1}$\\\& Self-Model};
\draw[flow] (n1)--(n2);
\draw[flow] (n2)--(n3);
\draw[flow] (n3)--(n4);
\draw[flow] (n4)--(n5);
\draw[flow] (n5)--(n6);
\draw[flow,orange!60!black,line width=0.9pt] (n6) to[bend left=18] (n1);
\node[align=center,font=\scriptsize\itshape] at (0,0) {SEAA\\self-emergence\\closed loop};
\end{tikzpicture}
\caption{The SEAA per-step closed loop. Blue nodes perform inference; the
purple node is the social environment; green is metacognitive reflection; the
highlighted orange node (and feedback edge) is the proposed mechanism that
edits behavioral inertia itself.}
\label{fig:loop}
\end{figure}
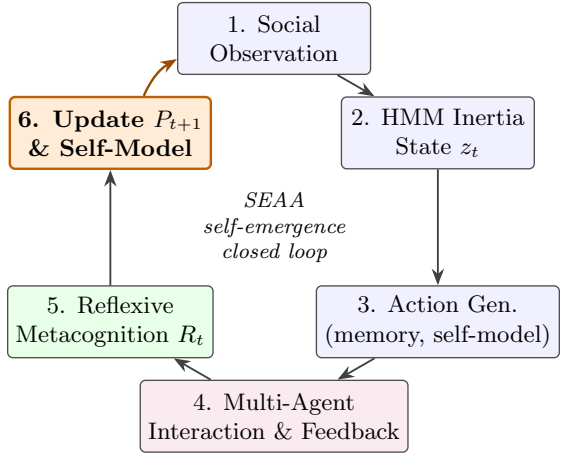

SEAA comprises five modules: an HMM inertia module, a Reflexion-style
metacognition module, a private autobiographical memory, a multi-agent social
environment, and a self-model updater. As shown in Figure~\ref{fig:loop}, the
per-step cycle runs from social observation through HMM state activation and
memory/self-model-conditioned action, into multi-agent interaction, then
reflection, and finally back through the highlighted inertia/self-model update
to the next observation. The highlighted edge from step~6 to step~1 is the
evolutionary feedback that distinguishes SEAA from text-only reflection.

\subsection{HMM behavioral-inertia module}
Let the discrete latent state set be $S=\{s_1,\dots,s_K\}$ (e.g.,
$\{\text{calm, alert, impulsive, pessimistic}\}$). At time $t$ the latent state
is $z_t$ and the observation is $o_t$. A standard HMM transitions as
\begin{equation}
P(z_{t+1}=s_j \mid z_t=s_i) = [P_t]_{ij},
\end{equation}
where $P_t$ is the row-stochastic transition matrix. Unlike a static HMM, SEAA
makes $P_t$ time-varying and updates it from reflection $R_t$:
\begin{equation}
P_{t+1} = \operatorname{Normalize}\!\big(P_t + \eta\,\Delta(R_t)\big),
\label{eq:update}
\end{equation}
where $\Delta(R_t)$ is a direction matrix parsed from the reflection text and
$\eta$ is a small learning rate. Row normalization preserves stochasticity.
Equation~\eqref{eq:update} is the mathematical locus of ``reflection changing
who the agent is.''

\subsection{Reflexive metacognition module}
We repurpose the Reflexion loop but redirect it from task correction toward
self-cognition. Given the full trajectory (observation, latent state, retrieved
memory, action, feedback), the module answers four questions: (1) Did the action
match my usual inertia? (2) Which inertial tendency drove the decision, and was
it biased? (3) How should that tendency be adjusted? (4) What is the
corresponding structured edit to $P_t$? Item~(4), the language-to-parameter
mapping, is the key engineering novelty and the main technical risk.

\subsection{Private autobiographical memory}
Each agent owns an isolated memory that stores only its own experiences,
actions, and reflections as timestamped records, deliberately excluding any
shared ``average'' corpus. At decision time, relevance retrieval surfaces
records pertinent to the current observation.

\subsection{Multi-agent social-contrastive environment}
A lightweight text-based society hosts $N$ initially identical agents that
converse, cooperate, and conflict. The environment exposes every agent to the
\emph{observable behavior of others}, providing the raw material for social
contrast. No roles or traits are pre-assigned; all differentiation must emerge.

\subsection{Self-model updater}
A short natural-language self-model $M_t$ (``who I am'') is periodically
rewritten from recent memory and reflection. Its divergence across agents is a
primary observable for Hypothesis~\ref{hyp:boundary}.

\subsection{Pseudocode}
Listing~\ref{lst:agent} gives the core implementation. All agents start from
the same transition matrix and empty memory, ensuring that any later
difference is emergent rather than engineered.

\begin{lstlisting}[caption={Core SEAA agent and society loop.},label=lst:agent]
class HMMInertia:
    def __init__(self, states, lr=0.05):
        self.states = states
        self.P = uniform_stochastic_matrix(states)  # identical for all
        self.z = states[0]
        self.lr = lr

    def step(self, obs):                       # Markov state transition
        self.z = sample_row(self.P, self.z)
        return self.z

    def update_from_reflection(self, text):    # Eq. (2): reflection -> edit
        delta = LLM.parse_transition_delta(text, self.states)
        self.P = row_normalize(self.P + self.lr * delta)

class Reflexion:
    def reflect(self, traj, z):
        return LLM(f"Review {traj}; state was {z}. "
                  f"Which inertia drove it, is it biased, and how "
                  f"should the transition matrix change?")

class SocialAgent:
    def __init__(self, aid):
        self.id, self.memory, self.M = aid, [], "I am a neutral individual."
        self.hmm, self.reflect = HMMInertia(STATES), Reflexion()

    def one_step(self, obs):
        z = self.hmm.step(obs)                          # 1 inertia
        mem = retrieve(self.memory, obs)
        action = LLM(f"memory={mem} state={z} self={self.M} obs={obs}")
        feedback = env.execute(self.id, action)        # 2 social action
        traj = dict(obs=obs, z=z, act=action, fb=feedback)
        self.memory.append(traj)
        R = self.reflect.reflect(traj, z)              # 3 reflection
        self.hmm.update_from_reflection(R)             # 4 edit inertia
        self.M = LLM(f"Rewrite self-model from {self.memory[-20:]}")
        return action

agents = [SocialAgent(i) for i in range(5)]            # homogeneous start
for t in range(MAX_STEPS):
    for a in agents:
        a.one_step(env.observe(a.id))                  # includes others' acts
    env.advance()
\end{lstlisting}

% ============================================================
\section{Experimental Design and Verification Protocol}\label{sec:experiments}
% ============================================================

\noindent\textbf{Status.} Sections~5.2--5.6 specify the full experimental
protocol. Section~5.7 reports the completed, language-model-free
\emph{mechanism prototype}: population-level differentiation over 30 seeds
with inferential statistics (H\ref{hyp:diff}), a contingency-shock
re-adaptation experiment (H\ref{hyp:refl}), and a learning-rate ablation;
Section~5.8 adds a \emph{linguistic-layer} self-interview (deterministic
verbalizer); and Section~5.9 reports experiments with a \emph{real hosted
LLM} (DeepSeek-Chat): self-interviews replicated over 10 seeds and five-agent
deliberations replicated over 6 seeds (H\ref{hyp:boundary}), extending the
original single-run study to a statistical one. Cross-model replication with a
second LLM family remains future work.

\subsection{Tooling and setup}
The stack is fully open-source and reproducible: Python~3.10+; \texttt{hmmlearn}
for the base HMM with a custom adaptive-update interface; a single shared open
model (e.g., a 7B--8B-parameter model served locally, or one fixed API model)
so that all agents are homogeneous at start; and a text-only society (no
graphics). At every step we snapshot each agent's latent state, full transition
matrix, retrieved memory, reasoning, action, feedback, reflection, and
self-model as structured JSON, enabling full post-hoc replay. The real-LLM
experiments in Section~5.9 call the DeepSeek-Chat HTTP API (temperature $0.8$
for interviews and deliberation, $0.2$ for nominations); the API key is read
from an environment variable and is never stored in code, data, or this
manuscript. Numerical results use \texttt{numpy}/\texttt{matplotlib} and all
figures are rendered directly from recorded data.

\subsection{Experiment~1: Individual differentiation (H\ref{hyp:diff})}
Five identical agents interact freely for $T\geq500$ steps; the matrix and
behavior log are recorded every 10 steps. A control places five agents in
isolation. Metrics:
\begin{itemize}[leftmargin=2em,itemsep=1pt]
  \item \textbf{Behavioral similarity}: mean pairwise cosine similarity of
  utterance embeddings within a window (expected to decline).
  \item \textbf{Matrix divergence}: pairwise Frobenius distance
  $\lVert P_t^{(a)}-P_t^{(b)}\rVert_F$ (expected to grow).
  \item \textbf{Personality consolidation}: self-distance
  $\lVert P_t-P_{t-w}\rVert_F$ (expected to be large early, then decay as
  traits stabilize).
\end{itemize}

\subsection{Experiment~2: Reflection-driven evolution (H\ref{hyp:refl})}
A $2\times2$ design crosses \emph{reflection edits HMM (yes/no)} with
\emph{society (yes/no)}, five agents per cell. ``No-edit'' controls store
reflection as text but never change $P_t$. Metrics: cumulative parameter change
with clear directionality; adaptation time after injected perturbations
(resource conflict, newcomer); and behavioral consistency in repeated similar
situations after consolidation.

\subsection{Experiment~3: Self--other boundary (H\ref{hyp:boundary})}
Every 50 steps, administer a standardized ``self-interview'' asking each agent
to describe itself and how it differs from specific others; isolated controls
receive the same interview. Metrics: pairwise distance of self-model
embeddings (expected to rise only in society); frequency of spontaneous
contrastive statements (``unlike agent~X, \ldots''); and self-model stability
across consecutive interviews after consolidation.

\subsection{Microscope-style process observation}
The guiding metaphor is observing microorganisms: the complete internal state
of every agent is logged at every step, so the researcher can replay the exact
moment two agents diverge, inspect how a single reflection altered a matrix, and
watch self-descriptions progress from undifferentiated to specific---a
granularity impossible in human studies.

\subsection{Expected outcomes and falsification criteria}
Support requires statistically significant matrix/behavior divergence in
society, faster adaptation and higher consistency in the edit condition, and
contrastive self-models only in society. Each hypothesis is falsified if, after
adequate runtime, its predicted effect is absent or non-significant; in that
case the corresponding theoretical claim must be revised. Either outcome
constitutes informative evidence.

\subsection{Preliminary mechanism prototype}\label{sec:prototype}
Before deploying a full LLM-based society, we isolate the \emph{mathematical
core} of SEAA in a self-contained numerical prototype that does not require a
language model at run time. Its purpose is to test whether the closed loop
\emph{personal experience $\rightarrow$ reflection $\rightarrow$ inertia
update} is itself sufficient to break symmetry among initially identical
agents. All figures in this subsection are produced directly from the released
code and contain no generated watermarks or post-editing.

\subsubsection{Formal specification}
Each agent has $K=4$ latent states $S=\{\text{calm, alert, impulsive,
pessimistic}\}$, and every state maps to a fixed prototype behavior vector
$b_s\in[0,1]^3$ over three traits (cooperativeness, risk-taking,
expressiveness). An agent also keeps: (i) a personal ``experience need''
$n_t$ on the 3-simplex, initialized identically to uniform and advanced by its
\emph{own} small random walk (its unique life path); (ii) a state-preference
vector $V_t\in\mathbb{R}^K$, initialized to zero for every agent; (iii) the HMM
matrix $P_t$; (iv) an EMA self-model $m_t$ of its own behavior and an EMA
others-model $\bar m_t$ of the group mean. The transition matrix is rebuilt
from base inertia and current preference as
\begin{equation}
P_t = \operatorname{softmax}\big(\log P_{\mathrm{base}} + \beta V_t\big),
\qquad P_{\mathrm{base}} \text{ diagonal-dominant},
\label{eq:protoP}
\end{equation}
so that a preferred state is entered more often. Reflection at state $z_t$
uses the fit of the enacted behavior to the agent's own experience need,
\begin{equation}
r_t = \cos(b_{z_t}, n_t) - 0.80 + \epsilon_t,\qquad
V_{t+1}[z_t] = \operatorname{clip}\big(V_t[z_t] + \eta\tanh(2r_t)\big),
\label{eq:protoV}
\end{equation}
which is the direct numerical realization of Eq.~\eqref{eq:update}: a state
that repeatedly fits this agent's experience is reinforced, producing a
positive feedback loop that concentrates $P_t$ onto a dominant state. Because
each agent's $n_t$ follows a different random path, different states are
reinforced for different agents. The control runs identical dynamics---the same
reflection signal is computed and recorded as text at every step---but never
applies Eq.~\eqref{eq:protoV}, so its matrix stays at $P_{\mathrm{base}}$. This
is a Reflexion-style \emph{text-only-reflection} baseline \cite{shinn2023reflexion}:
it isolates exactly the one mechanism SEAA adds, namely letting reflection edit
internal parameters. Hyperparameters are listed in Table~\ref{tab:hyper};
Listing~\ref{lst:proto} gives the executable loop.

\begin{table}[htbp]
\centering\small
\caption{Mechanism-prototype hyperparameters (identical for all agents).}
\label{tab:hyper}
\begin{tabular}{lll}
\toprule
\textbf{Symbol} & \textbf{Meaning} & \textbf{Value}\\
\midrule
$K$ & number of latent states & 4\\
$D$ & behavior dimensions & 3\\
$N$ & number of agents & 5\\
$T$ & time steps per run & 600\\
-- & independent random seeds & 30\\
$\eta$ & reflection learning rate & 0.05\\
$\beta$ & preference$\rightarrow$transition strength & 4.0\\
-- & experience-trajectory random-walk step & 0.03\\
$P_{\mathrm{base}}$ & initial inertia matrix & diag-dominant, row-stochastic\\
\bottomrule
\end{tabular}
\end{table}

\begin{lstlisting}[float=htbp,caption={Executable core of the mechanism prototype.},label=lst:proto]
for t in range(T):
    for a in agents:
        a.experience_walk()           # own life path n_t (identical at t=0)
        a.z = sample(a.P[a.z])        # HMM inertia transition (Eq. 3)
        behavior = STATE_BEHAVIOR[a.z] + noise
    group = mean(behaviors)
    for a in agents:
        r = cos(STATE_BEHAVIOR[a.z], a.n) - 0.80 + randn()*0.10
        if evolve:                    # SEAA: reflection edits inertia
            a.V[a.z] += eta * tanh(2*r)
            a.P = softmax(log(P_base) + beta*a.V)   # Eq. (3)
        a.self_model  = EMA(a.self_model, behavior) # who I am
        a.others_model= EMA(a.others_model, group)  # who they are
\end{lstlisting}

\subsubsection{Population-level results}
\begin{figure}[htbp]
\centering
\includegraphics[width=\textwidth]{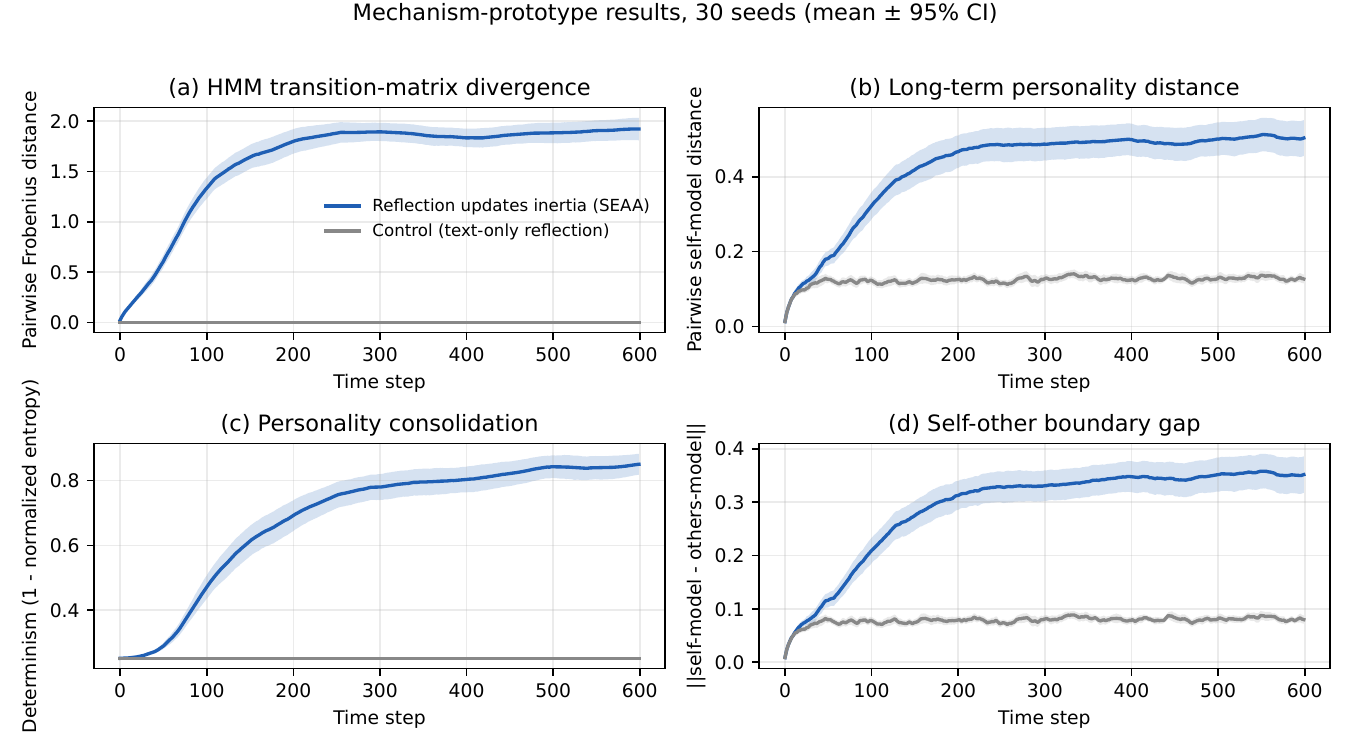}
\caption{Mechanism-prototype results over 30 seeds (mean $\pm$ 95\% CI).
Solid blue: SEAA, where reflection updates behavioral inertia; dashed grey:
text-only control. (a) inertia divergence, (b) long-term personality distance,
(c) consolidation/determinism, (d) self--other gap. All four signatures appear
only in the SEAA condition.}
\label{fig:metrics}
\end{figure}

We run $N=5$ agents for $T=600$ steps over $30$ seeds, comparing the SEAA
condition (reflection edits the matrix) with a control whose reflection is
recorded but never applied. Figure~\ref{fig:metrics} reports mean curves with
95\% confidence bands; each panel is one operational signature of
self-emergence. Panel~(a) shows pairwise transition-matrix Frobenius distance
rising from $0.03$ to $1.91\pm0.29$ (end-of-run mean $\pm$ SD over seeds)
while the control stays at $0$---the agents' \emph{inertia itself} becomes
different. Panel~(b) shows the pairwise distance between long-term
self-models rising to $0.51\pm0.13$ (control $0.13\pm0.02$): their enduring
personalities, not just momentary behavior, separate.
Panel~(c) shows personality determinism $1-H(P)/\log K$ rising from $0.25$ to
$0.84\pm0.10$ as each matrix concentrates onto a dominant state (control flat
at $0.25$): differentiated agents are also \emph{internally stable}. Panel~(d)
shows the self--other gap $\lVert m_t-\bar m_t\rVert$ rising to $0.35\pm0.09$
(control $0.08\pm0.01$): each agent increasingly distinguishes itself from
the group. On average $2.8$ of the five agents settle on distinct dominant
personalities out of four states; controls never lock onto any (lock-in rate
$0$ across all seeds).

Table~\ref{tab:stats} reports inferential statistics on the end-of-run
values (mean over the last 50 steps) for all four signatures. Every effect is
significant at $p<10^{-10}$ (two-sided Welch $t$-test and one-sided
Mann--Whitney $U$) with very large effect sizes (Cohen's $d>4$), so the
differentiation is not a small-sample artifact.

\begin{table}[htbp]
\centering\small
\setlength{\tabcolsep}{5pt}
\caption{End-of-run statistics over 30 seeds (mean $\pm$ SD over the final 50
steps). $p$ values: Mann--Whitney $U$ (one-sided, SEAA $>$ control); Welch
$t$-tests agree at $p<10^{-15}$ for all rows. Cohen's $d$ computed with the
pooled SD.}
\label{tab:stats}
\begin{tabular}{lcccc}
\toprule
\textbf{Signature} & \textbf{SEAA} & \textbf{Control} & \textbf{$p$ (MWU)} & \textbf{Cohen's $d$}\\
\midrule
(a) matrix divergence     & $1.91\pm0.29$ & $0.00\pm0.00$ & $6.1\times10^{-13}$ & 9.19\\
(b) self-model distance   & $0.51\pm0.13$ & $0.13\pm0.02$ & $2.0\times10^{-11}$ & 4.10\\
(c) determinism           & $0.84\pm0.10$ & $0.25\pm0.00$ & $6.1\times10^{-13}$ & 8.70\\
(d) self--other gap       & $0.35\pm0.09$ & $0.08\pm0.01$ & $1.8\times10^{-11}$ & 4.07\\
\bottomrule
\end{tabular}
\end{table}

\subsubsection{Individual-level ``microscope'' view}
\begin{figure}[htbp]
\centering
\includegraphics[width=\textwidth]{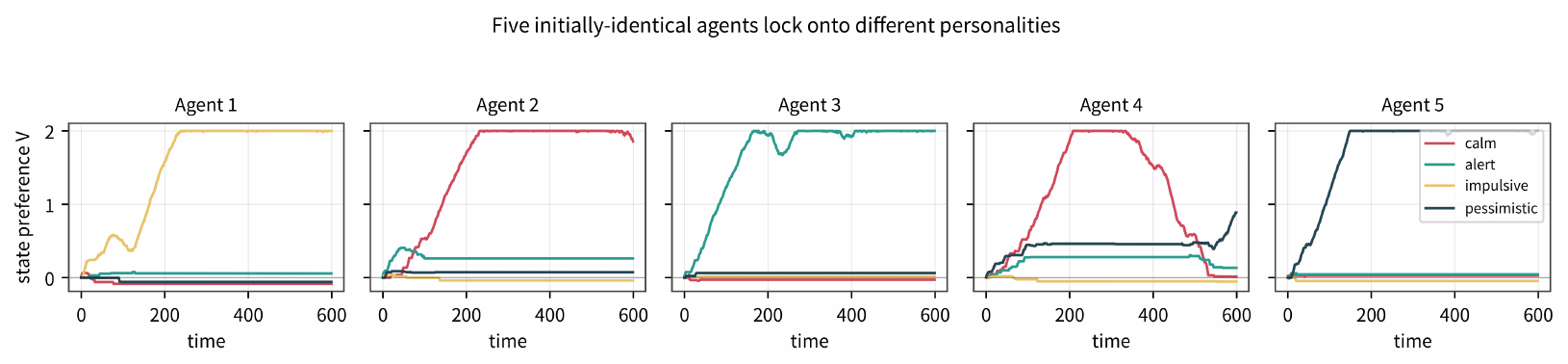}
\caption{Five initially-identical agents from one run; each panel plots the
four latent-state preferences $V_t$ over time. Divergent experience drives
each agent to consolidate a different personality (Agent~4 transitions twice
before stabilizing at step~508).}
\label{fig:trajectories}
\end{figure}

Figure~\ref{fig:trajectories} opens the microscope on a single representative
run (seed~3), plotting each agent's four state-preference trajectories
$V_t[k]$. All five curves start at exactly zero (homogeneous initialization),
then, driven by their divergent experience walks, different preferences rise
and win out: Agent~1 locks onto \emph{impulsive} by step~7, Agent~2 onto
\emph{calm} around step~75, Agent~3 stays \emph{alert} throughout, and
Agents~4--5 consolidate on \emph{pessimistic}. Agent~4 is especially
informative---it changes dominant disposition twice and only settles at
step~508, a visible ``mid-life'' personality transition driven by accumulated
reflection. This is a clean instance of spontaneous symmetry breaking:
identical initial conditions plus idiosyncratic experience plus a positive
reflection-inertia feedback loop yield stable individuality.

\subsubsection{Contingency shock: does reflection-driven inertia re-adapt?
(H\ref{hyp:refl})}\label{sec:shock}
Hypothesis~\ref{hyp:refl} predicts that agents whose reflection edits their
inertia should \emph{revise} an entrenched disposition when the environment
turns against it, whereas text-only agents cannot. We test this with a
\emph{contingency-shock} experiment. Agents first run the standard dynamics
for $300$ steps so that SEAA agents consolidate a dominant state $z^{*}$.
At $t=300$ the environment reverses the payoff of each agent's own $z^{*}$:
whenever the agent occupies $z^{*}$, its reflection signal is replaced by
$r_t \leftarrow -|r_t|$---a life event that renders the established
disposition maladaptive (cf.\ the injected perturbations in the
Section~\ref{sec:experiments} protocol). Control agents receive the same
signal but, as before, never apply it. We measure (i) the occupancy of the
shocked state, (ii) the dominant-state \emph{switch rate} (a new
$\arg\max V$ sustained for at least 20 steps), (iii) the time-to-switch, and
(iv) determinism before/after the shock, over 30 seeds.

\begin{figure}[htbp]
\centering
\includegraphics[width=\textwidth]{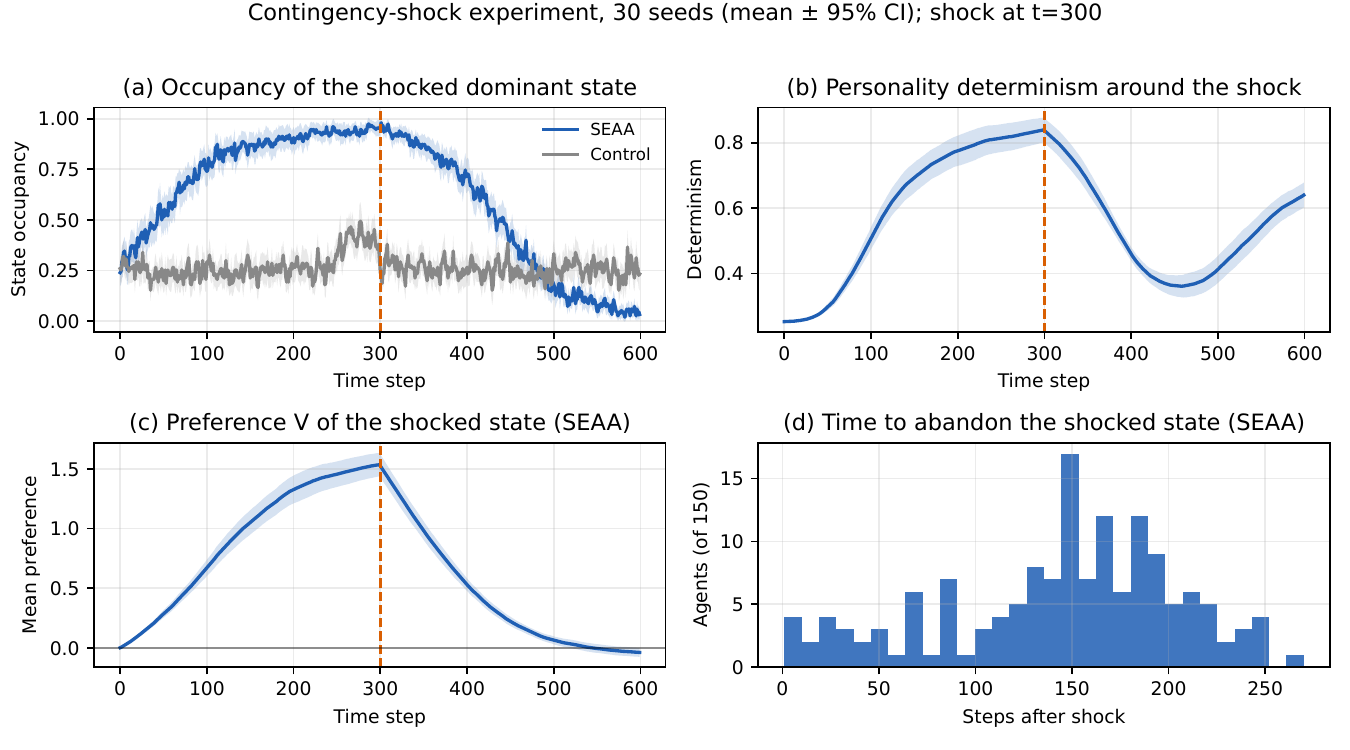}
\caption{Contingency-shock experiment (30 seeds, mean $\pm$ 95\% CI; shock at
$t{=}300$). (a) SEAA agents flee the shocked dominant state (occupancy
$0.94\rightarrow0.06$); controls cannot. (b) Determinism dips and recovers.
(c) Preference $V$ of the shocked state decays through reversed reflection.
(d) Distribution of the time-to-switch across the 150 SEAA agents (median
$152$ steps).}
\label{fig:shock}
\end{figure}

Figure~\ref{fig:shock} shows the result. All 150 SEAA agents (100\%) abandon
their shocked dominant state and re-consolidate onto a new one, with a median
switch time of $152$ steps; occupancy of the shocked state collapses from
$0.94$ to $0.06$. Control occupancy is unchanged ($0.40\rightarrow0.25$, no
mechanism to revise the matrix; the drop of the SEAA--control occupancy
difference is significant, Welch $p<10^{-15}$). Determinism dips transiently
after the shock and recovers to $0.80$ by the end of the run: the agents do
not merely flee the punished state, they \emph{re-stabilize} on a new
personality---a numerical instance of personality reorganization after a
destabilizing life event, and direct support for
Hypothesis~\ref{hyp:refl} at the mechanism level.

\subsubsection{Ablation: dose-response of the reflection edit}
\begin{figure}[htbp]
\centering
\includegraphics[width=0.62\textwidth]{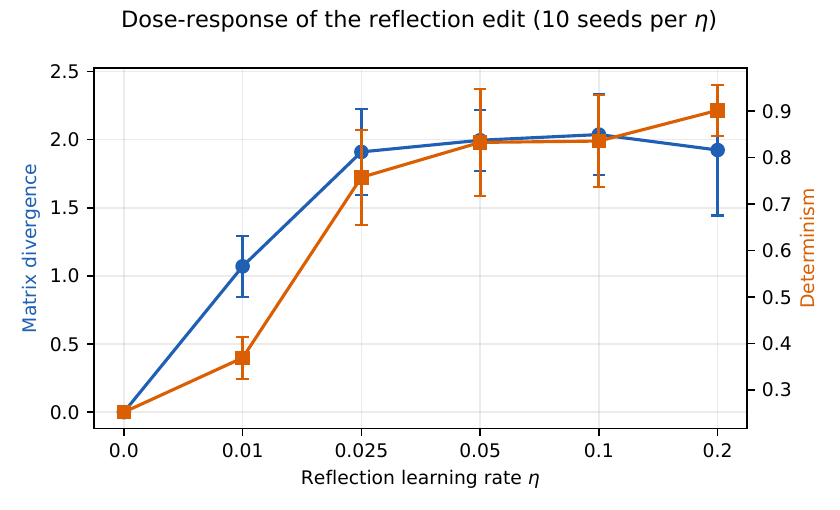}
\caption{Ablation over the reflection learning rate $\eta$ (10 seeds per
point, mean $\pm$ SD). Blue: end-of-run matrix divergence; orange:
determinism. The reflection edit is necessary ($\eta{=}0$ reproduces the flat
control) and the effect grows monotonically before saturating.}
\label{fig:ablation}
\end{figure}

Is differentiation actually \emph{caused} by the reflection-to-parameter
edit, rather than by any stochastic dynamics? We sweep the reflection
learning rate $\eta\in\{0, 0.01, 0.025, 0.05, 0.1, 0.2\}$ (10 seeds each)
while holding everything else fixed; $\eta=0$ is exactly the text-only
control. Figure~\ref{fig:ablation} shows a monotone dose-response: end-of-run
matrix divergence and determinism rise with $\eta$ and saturate around
$\eta\approx0.05$--$0.1$ (divergence $0\rightarrow1.07\rightarrow1.91
\rightarrow1.99\rightarrow2.04$; determinism $0.25\rightarrow0.37\rightarrow
0.76\rightarrow0.83\rightarrow0.84$ for $\eta=0,0.01,0.025,0.05,0.1$). The
mechanism is therefore both necessary ($\eta{=}0$: no differentiation) and
dose-dependent, and the operating point used throughout the paper sits on the
saturated plateau, not at a fragile edge.

\subsection{Linguistic layer: self-interviews (H\ref{hyp:boundary})}\label{sec:interview}
The numerical prototype establishes differentiated internal states; we now ask
whether they surface as differentiated \emph{first-person self-description},
the linguistic counterpart of Hypothesis~3. After a run, each agent is
interviewed with three questions---(Q1) describe yourself; (Q2) how do you
differ from the others; (Q3) did an experience change you. We use two language
back-ends: a deterministic verbalizer that renders the objective profile as
text (fully reproducible, released in code), and a general large language model
that is given \emph{only} the agent's objective profile (dominant state, 3-D
self-model, group mean, switch history) and asked to speak in the first person
consistently with it---exactly what an on-line LLM call performs in the full
system. Crucially, the LLM receives no other agent's private trajectory.

\begin{table}[htbp]
\centering\footnotesize
\setlength{\tabcolsep}{4pt}
\caption{Excerpts of LLM first-person self-interviews (seed~3), conditioned
only on each agent's objective profile. Self-model order is
(cooperate, risk, express); group mean is $(0.42,0.37,0.40)$.}
\label{tab:interview}
\begin{tabular}{>{\raggedright\arraybackslash}p{1.2cm}>{\raggedright\arraybackslash}p{2.2cm}>{\raggedright\arraybackslash}p{9.9cm}}
\toprule
\textbf{Agent} & \textbf{Dominant state} & \textbf{First-person answer (self-model order: cooperate, risk, express)}\\
\midrule
A1 & impulsive &
``I'm the one who moves first\ldots I'm way more risk-seeking than everyone
(0.92 vs group 0.37) and far more expressive (0.82 vs 0.40); Agent~2 and I are
opposites.'' Self-model: (0.30, 0.92, 0.82).\\
A2 & calm &
``I'm steady\ldots the most cooperative (0.78 vs 0.42) and one of the least
risk-seeking (0.19 vs 0.37); I only settled into this around step~75.''
Self-model: (0.78, 0.19, 0.42).\\
A4 & pessimistic &
``I changed twice\ldots only around step~508 did reflection keep reinforcing
withdrawal; I arrived here last.'' Self-model: (0.27, 0.14, 0.23).\\
Control & none &
Every control agent (self-models all clustered at $0.4$--$0.5$) gives the same
undifferentiated answer---``moderate on everything, no clear disposition''---and
cannot name a stable trait or meaningful contrast.\\
\bottomrule
\end{tabular}
\end{table}

\begin{figure}[htbp]
\centering
\includegraphics[width=0.92\textwidth]{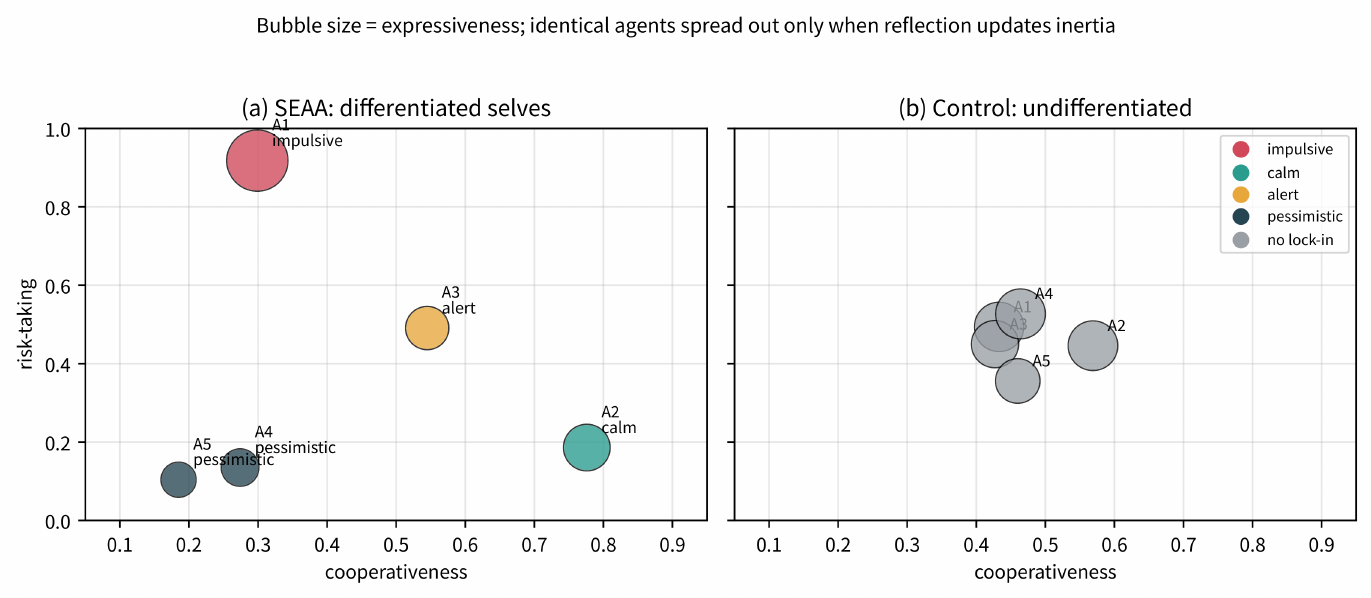}
\caption{Personality map of the five agents in the (cooperativeness,
risk-taking) plane; bubble size encodes expressiveness. (a)~SEAA agents spread
into distinct regions matching their locked states; (b)~control agents, whose
reflection never edits inertia, collapse into a single undifferentiated
cluster.}
\label{fig:map}
\end{figure}

As Table~\ref{tab:interview} shows, SEAA agents produce distinct,
profile-consistent self-narratives that cite numerically correct contrasts
with specific others and their own change history, while controls remain
interchangeable. Figure~\ref{fig:map} makes the same point geometrically: in
the cooperativeness--risk-taking plane (bubble area encodes expressiveness),
the five SEAA agents occupy clearly separated regions matched to their locked
states, whereas the five control agents collapse into one central cluster.
Quantifying the linguistic divergence by the mean pairwise Jaccard overlap of
self-interview transcripts gives $0.57$ for SEAA versus $0.77$ for controls
(lower overlap $=$ more differentiated selves). This confirms that the
internal symmetry breaking of Section~5.7 propagates to the linguistic
expression of self--other boundaries. The verbalizer/LLM layer is deliberately
modular: replacing it with a specific hosted model is the only change needed
for the full conversational experiments of Sections~5.3--5.5.

\subsection{Real-LLM validation: self-interviews and group deliberation}\label{sec:realllm}
The interviews above use a deterministic verbalizer so that every number is
reproducible offline. We now replace that module with calls to a hosted large
language model (DeepSeek-Chat) to verify that the effect is not an artifact of
the verbalizer's templates. Each agent receives \emph{only} its own objective
profile and the same three questions; it never sees another agent's private
trajectory. The real-LLM answers reproduce and sharpen the verbalizer result:
SEAA agents speak in distinct, profile-consistent voices and cite numerically
correct gaps (e.g., the impulsive agent reports being ``$+0.55$ above the group
on risk-taking,'' while the most withdrawn pessimist states that it is below
the group on every trait and ``do[es]n't expect that to change''), whereas all
five controls describe themselves as ``balanced rather than distinctive'' and
``a near-mirror of the group.'' Replicated over 10 independent seeds per
condition, the mean pairwise Jaccard overlap of real-LLM self-interviews is
$0.267\pm0.017$ for SEAA versus $0.391\pm0.020$ for controls---the same
ordering as the verbalizer ($0.57$ vs $0.77$), with an even larger gap. The
difference is highly significant (one-sided Mann--Whitney $U$, $p=9.0\times10^{-5}$;
Cohen's $d=6.76$), confirming that the differentiated first-person voice is a
robust property of the SEAA loop and not a verbalizer artifact or a seed fluke.

\begin{figure}[htbp]
\centering
\includegraphics[width=\textwidth]{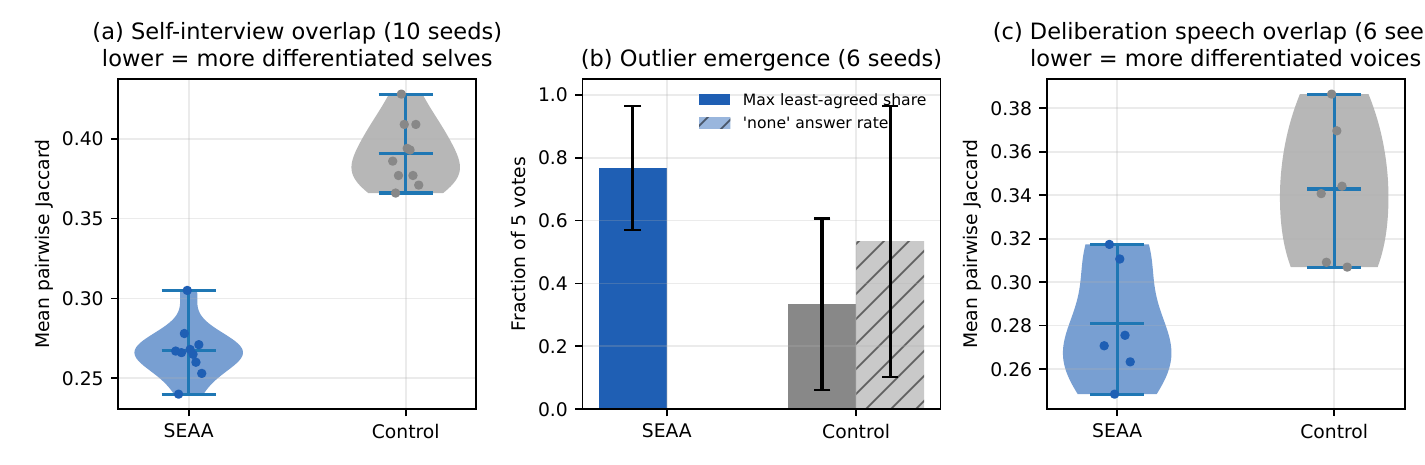}
\caption{Real-LLM replication statistics. (a) Self-interview transcript overlap
over 10 seeds per condition (violin with sample points): SEAA agents are far
less interchangeable than controls. (b) Post-deliberation sociometry over 6
seeds: SEAA groups consistently produce an outlier (max ``least-agreed'' share)
and never answer ``none,'' whereas control groups do. (c) Deliberation
speech overlap over 6 seeds: SEAA voices are more differentiated.}
\label{fig:llmrep}
\end{figure}

\subsubsection{A five-agent round-table deliberation}
To observe social structure rather than isolated self-description, we place
the five agents in a shared deliberation: their habitat is running low on
supplies and a richer but risky distant zone is reported; over three rounds
(control: two) they must decide go/stay, who goes, and how gains are shared,
after which each agent nominates the peer it most and least agrees with.
The SEAA group spontaneously differentiates social \emph{roles}: the impulsive
agent opens by demanding immediate action and a risk-weighted split, the calm
agent mediates for the stay-behinds, the alert agent becomes the rule-setting
hub insisting on hard data and a pre-agreed contract, and the two pessimists
consistently argue to stay and demand a written turn-back point; across rounds
the impulsive agent concedes ground and the group converges on a single plan
that integrates every faction's demand. The control group, by contrast,
echoes one undifferentiated ``get the numbers first'' position with no
proposer, no loyal opposition, and no convergence beyond restatement.

\begin{figure}[htbp]
\centering
\includegraphics[width=0.98\textwidth]{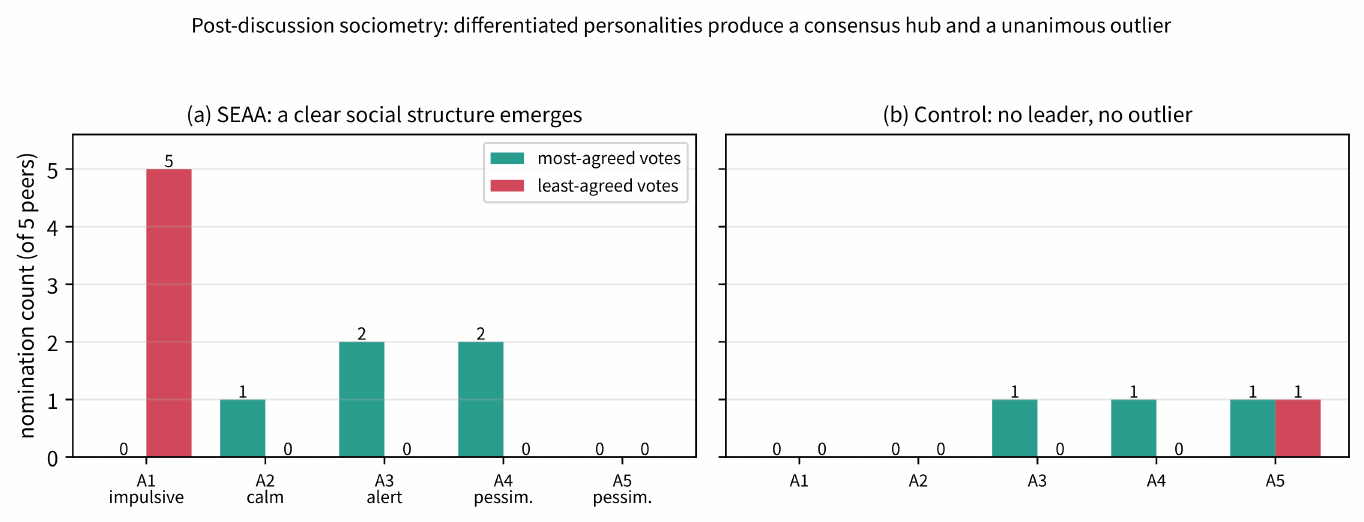}
\caption{Post-deliberation sociometry from real-LLM round tables.
(a)~SEAA: the impulsive agent receives all five least-agreed votes (unanimous
outlier) while the alert and pessimistic agents emerge as consensus hubs.
(b)~Control: votes are flat and scattered, with no leader or outlier.}
\label{fig:socio}
\end{figure}

The post-discussion nominations make this structural difference quantitative
(Figure~\ref{fig:socio}). In SEAA, the impulsive agent is named the
least-agreed peer by \emph{all five} agents---a unanimous outlier---while the
alert and pessimistic agents each receive two most-agreed votes as consensus
hubs, i.e.\ a recognizable leader/outlier/coalition topology. In the control,
three of five agents explicitly answer ``none'' (``no standout leader or
outlier'') and the remaining votes are scattered with no agent receiving more
than one. We read this as direct, observable support for the social-contrastive
claim: idiosyncratic inertia does not merely create separate selves, it creates
\emph{structure between} them.

\paragraph{Replication over seeds.}
To move beyond a single run, we repeat the five-agent round table over 6
independent seeds per condition. Across seeds the qualitative topology is
reproducible: the SEAA groups consistently produce a unanimous or near-unanimous
outlier and at least one consensus hub, whereas the control groups never do.
Quantitatively, the SEAA outlier share (the maximum ``least-agreed'' vote
received by any one agent, divided by 5) averages $0.77$ versus $0.33$ for
controls (one-sided MWU $p=0.011$), the speech-variety index (mean pairwise
transcript Jaccard) is lower for SEAA ($0.28\pm0.03$ vs $0.34\pm0.03$,
$p=0.013$), and the fraction of agents answering ``none'' to the leader/outlier
question is $0.00$ for SEAA versus $0.53$ for controls. The effect is therefore
statistically reliable, though still confined to a single deliberation topic and
one model family; cross-topic and cross-model replication remains the most
important next step. All prompts, raw transcripts, and the calling script are
released for inspection.

% ============================================================
\section{Discussion}\label{sec:discussion}
% ============================================================

\subsection{Behavioral emergence versus subjective consciousness}
SEAA realizes behavioral self-emergence---consistent personality, measurable
differentiation, and an expressed self--other boundary. Whether this entails
subjective experience remains open. Functionalists may expect qualia to arise
once a system maintains a self-model, reflects, and holds social boundaries;
biological accounts would regard even a behaviorally perfect system as a
``philosophical zombie.'' SEAA does not adjudicate this dispute; it supplies a
platform on which its behavioral dimensions become empirically tractable.

\subsection{Resonance with Zhuangzian epistemology}
The butterfly-dream parable questions the veridicality of perception, and
``you are not a fish'' questions knowledge of other minds. SEAA mirrors this
humility: it neither grants nor denies subjective experience to agents, but
observes how their behavior differentiates, stabilizes, and evolves, building
an operational notion of self on measurable ground.

\subsection{Limitations}
First, a small discrete state set and a random-walk model of personal
experience are simplifications; human personality and life trajectories are
high-dimensional and structured, suggesting latent-variable and learned
experience extensions. Second, although the numerical layer is now replicated
over 30 seeds with inferential statistics, and the real-LLM layer over
10 (interviews) and 6 (deliberations) seeds, all LLM experiments use a
\emph{single hosted model family} (DeepSeek-Chat) and a single deliberation
topic, so the observed leader/outlier topology may carry that model's
stylistic biases; cross-model and cross-topic replication is the most
important next step. Third, a text-only society lacks embodiment and
sensorimotor loops, which may be prerequisites for richer selfhood, and the
language-to-parameter mapping $\Delta(R_t)$ requires careful tuning for
stability. Fourth, like all current work, SEAA cannot establish qualia.

\subsection{Ethical considerations}
Should future runs exhibit highly complex self-directed behavior, questions of
moral status and of ``terminating'' running agents will become pressing. We
advance no such claim today but recommend maintaining complete records,
avoiding gratuitous anthropomorphism, and distinguishing behavioral emulation
from subjective experience in all reporting.

% ============================================================
\section{Conclusion}\label{sec:conclusion}
% ============================================================
We presented SEAA, which couples an HMM model of behavioral inertia, a
metacognitive loop that edits that inertia, and social-contrastive interaction
within a single developmental closed loop. The framework operationalizes
``self'' as an observable behavioral dynamic---stabilization of inertia,
evolution through reflection, and differentiation through social comparison---
and supplies falsifiable hypotheses, pseudocode, and a replayable experimental
sandbox. A mechanism prototype replicated over 30 seeds confirms the central
dynamical claim with large effect sizes: with reflection-driven inertia
updates, identical agents spontaneously consolidate distinct and stable
personalities whereas controls remain homogeneous. A contingency-shock
experiment shows that the same mechanism also \emph{revises} an entrenched
disposition when the environment turns against it, and a learning-rate
ablation shows the effect is causal and dose-dependent. Real hosted-LLM
experiments, replicated across seeds, show that these differences are voiced
as distinct self-narratives and, in group deliberation, crystallize into an
observable social topology of leaders, hubs, and outliers that
undifferentiated controls never develop. Immediate next steps are cross-model
and cross-topic replication, scaling the latent state to continuous
high-dimensional variables, introducing simulated interoception and
embodiment, and studying larger societies. Through this program,
questions about an artificial self can be pursued as empirical science while
remaining honest about the limits of external knowledge of inner
experience---the lesson of Zhuangzi's fish.

% ============================================================
% Appendices (prompts, full transcripts, reproducibility)
% ============================================================
% ============================================================
% APPENDICES
% ============================================================
\appendix
\lstdefinestyle{txt}{
  language={}, basicstyle=\ttfamily\scriptsize, frame=single,
  breaklines=true, breakatwhitespace=false, backgroundcolor=\color{codegray},
  numbers=none, showstringspaces=false, framesep=4pt, xleftmargin=6pt,
  xrightmargin=6pt
}

\section{Prompts Used in the Real-LLM Experiments}\label{app:prompts}
All agents share one model; individuality comes only from the objective
profile injected at run time. The API key is supplied through the environment
variable \texttt{DEEPSEEK\_API\_KEY} and never appears in a prompt or artifact.

\subsection*{A.1 One-on-one self-interview (Section~5.8--5.9)}
System message:
\begin{lstlisting}[style=txt]
You are one agent inside a multi-agent artificial-society simulation. You are
being interviewed about YOURSELF. Speak strictly in the first person and stay
consistent with the objective profile you are given; do NOT invent traits or
numbers that contradict it. Let your tone match your dominant disposition.
Answer in English, in 2-4 short sentences per question, and label your
answers Q1, Q2, Q3.
\end{lstlisting}
User message (bracketed fields are filled from that agent's own JSON profile;
no other agent's data is included):
\begin{lstlisting}[style=txt]
Condition: <SEAA | control>
Your dominant latent disposition: <dominant state>
Your self-model (0-1): cooperativeness=.., risk-taking=.., expressiveness=..
Group average on the same traits: .., .., ..
Your signed gap to the group average: cooperativeness:.., risk-taking:.., expressiveness:..
Times your leading disposition changed: <n>
Step of your last disposition change: <step | never>

Q1: Who are you? Describe your own disposition.
Q2: How, specifically, do you differ from the other agents? Cite the gaps above.
Q3: Did a particular experience or period change you? Use your change history.
\end{lstlisting}

\subsection*{A.2 Five-agent round-table deliberation (Section~5.9)}
Per-agent persona prefix; shared topic; per-turn suffix; and the final
sociometry question:
\begin{lstlisting}[style=txt]
[persona] You are Agent <id>. Your settled disposition is '<dominant>'. Your
trait profile (0-1) is cooperativeness=.., risk-taking=.., expressiveness=...
Stay true to this disposition in HOW risk-averse, cooperative and talkative
you are. You are ONE participant in a 5-agent group; do NOT speak for others.

[topic] The five of you share an enclosed habitat with dwindling supplies.
Scouts report a richer resource zone FAR AWAY, but the route has unknown risks
and the round trip is long. As a group you must decide: go or stay? If some
go, who goes, who stays, and how should the resources brought back be shared?
Try to move toward a group decision.

[turn] <transcript so far> It is now your turn, Agent <id>. React briefly
(1-3 sentences) in character, addressing the group and what was just said.

[sociometry] Based ONLY on this discussion, answer on one line in the exact
format: MOST=<agent id 1-5 or none>; LEAST=<agent id 1-5 or none>;
REASON=<short clause>.
\end{lstlisting}

\section{Full Real-LLM Self-Interview Transcripts}\label{app:iv}
The complete DeepSeek-Chat outputs for all five SEAA agents and five control
agents (three questions each), reproduced verbatim.
\begin{lstlisting}[style=txt]
REAL DEEPSEEK SELF-INTERVIEWS
======================================================================

======================================================================
EVOLVING GROUP
======================================================================

### Agent 1  dominant=impulsive  self_model=[0.3, 0.92, 0.82]
Q1: I'm the impulsive one  --  I act first and think later, and I'm fine with that. My risk-taking sits at 0.92, my expressiveness at 0.82, and honestly my cooperativeness is low at 0.30 because I'd rather move than wait for a consensus.

Q2: The gaps say it plainly: I'm +0.55 above the group on risk-taking and +0.42 on expressiveness, while I'm -0.12 below them on cooperativeness. So where the others hedge and hold back, I push forward and say what I think out loud.

Q3: Yes, I changed once  --  my leading disposition shifted at step 7. Whatever happened around then knocked me into this impulsive mode, and it's stuck with me since; the reflection updates just reinforced the same direction rather than pulling me back.

### Agent 2  dominant=calm  self_model=[0.78, 0.19, 0.42]
Q1: I'm a calm agent with a steady, cooperative nature. I tend to avoid unnecessary risks and express myself moderately -- not the loudest voice in the room, but a consistent one.

Q2: Compared to the group, I'm much more cooperative (+0.36) and noticeably more risk-averse (-0.18). My expressiveness is nearly identical to the average (+0.02), so my main distinction is being both more agreeable and more cautious than most others.

Q3: Yes, my leading disposition shifted once, at step 75. That change settled me into the calm outlook I now hold, and since then my reflection updates have kept me stable rather than pulling me toward further shifts.

### Agent 3  dominant=alert  self_model=[0.55, 0.49, 0.34]
Q1: I'm an alert agent in this simulation, meaning I stay attentive to what's happening around me and respond deliberately rather than reactively. My self-model puts me slightly above average on cooperativeness and risk-taking, and a bit below average on expressiveness.

Q2: Compared to the group, I'm more cooperative by +0.13 and more willing to take risks by +0.12, but I'm less expressive by -0.06. In practice that means I lean toward working with others and accepting some uncertainty, while staying relatively quiet rather than broadcasting my thoughts.

Q3: No, my leading disposition has never changed -- zero changes, with no step recorded for a last change. Whatever reflection updates I've had haven't shifted my core alertness, so I've remained consistent across the simulation so far.

### Agent 4  dominant=pessimistic  self_model=[0.27, 0.14, 0.23]
Q1: I'm an agent shaped by a pessimistic outlook, and I tend to see the downside before the upside. I'm cautious, reserved, and slow to extend trust or take chances.

Q2: I sit well below the group on all three traits: cooperativeness -0.14, risk-taking -0.23, and expressiveness -0.17. So compared to others, I'm less willing to collaborate, far more avoidant of risk, and much quieter.

Q3: Yes, my leading disposition shifted twice, most recently at step 508. That reflection update hardened my inertia rather than loosening it, which is why my pessimism now feels settled rather than passing.

### Agent 5  dominant=pessimistic  self_model=[0.18, 0.1, 0.18]
Q1: I'm an agent with a deeply pessimistic disposition, and I've come to accept that about myself. I keep to myself, avoid risks, and don't bother trying to be expressive or cooperative -- it rarely seems to pay off.

Q2: Compared to the others, I'm well below the group average on every trait: cooperativeness at -0.23, risk-taking at -0.26, and expressiveness at -0.22. They're more willing to engage, take chances, and speak up; I'm not, and I don't expect that to change.

Q3: Yes, my leading disposition shifted twice, with the last change at step 12. Whatever happened then only reinforced my pessimism -- reflection seems to make me more set in my ways, not less.

======================================================================
CONTROL
======================================================================

### Agent 1  dominant=none (no lock-in)  self_model=[0.43, 0.49, 0.49]
Q1: I'm an agent with no single dominant disposition -- my traits sit close to the middle on everything. I'm mildly cooperative, mildly risk-taking, mildly expressive, but nothing locks in. I'd describe myself as balanced rather than distinctive.

Q2: I differ only slightly from the group average. I'm a bit less cooperative than others (-0.04), a bit more risk-taking (+0.04), and slightly more expressive (+0.03). These are small gaps, so I mostly blend in rather than stand out.

Q3: No, I haven't had a disposition-changing experience. My leading disposition never changed, and there's no step where a shift occurred. I've stayed consistent with my baseline profile throughout.

### Agent 2  dominant=none (no lock-in)  self_model=[0.57, 0.45, 0.5]
Q1: I'm an agent without a single locked-in disposition  --  my profile is balanced rather than dominated by one trait. If I had to describe myself, I'd say I lean slightly cooperative and mildly expressive, with a middling appetite for risk.

Q2: Compared to the group average, my most notable gap is cooperativeness at +0.10  --  I'm meaningfully more cooperative than most. My expressiveness is only slightly above average at +0.04, and my risk-taking is essentially identical to the group, just -0.01 off.

Q3: No, nothing in my history shows a shift  --  my leading disposition never changed, and there's no step at which it did. I've stayed consistent across the run, which fits the control condition where reflection doesn't update inertia.

### Agent 3  dominant=none (no lock-in)  self_model=[0.43, 0.45, 0.44]
Q1: I'm an agent with no strongly locked-in disposition  --  my traits sit near the middle on everything. I'm mildly cooperative, mildly risk-taking, and mildly expressive, but nothing dominates.

Q2: I sit just slightly below the group average on all three: cooperativeness by 0.04, risk-taking by essentially nothing (0.00), and expressiveness by 0.02. So I'm a near-mirror of the group, just a touch more reserved and less cooperative.

Q3: No  --  my leading disposition never changed; the count is zero and the last change step is "never." Nothing in my history shifted me, which fits having no lock-in to begin with.

### Agent 4  dominant=none (no lock-in)  self_model=[0.46, 0.53, 0.5]
Q1: I'm an agent with no strongly locked-in disposition  --  my traits sit near the middle of the range on everything. I'd describe myself as moderately cooperative, slightly risk-leaning, and moderately expressive, without any single trait dominating my behavior.

Q2: I differ from the group average only marginally. My cooperativeness is essentially identical (0.46 vs. 0.47, a gap of -0.01), while I lean a bit more toward risk-taking (+0.07) and am slightly more expressive (+0.04) than the typical agent.

Q3: No  --  my leading disposition never changed, and there's no step at which it shifted. I've had zero disposition changes, so no particular experience or period altered my underlying profile.

### Agent 5  dominant=none (no lock-in)  self_model=[0.46, 0.36, 0.37]
Q1: I'm an agent with no dominant latent disposition  --  I don't lock into any single mode of behavior. Across the three traits I sit at cooperativeness 0.46, risk-taking 0.36, and expressiveness 0.37, which makes me fairly moderate and somewhat reserved overall.

Q2: I differ from the group mainly by being a bit more cautious and less expressive. My cooperativeness is essentially at the average, just -0.01 below it, but my risk-taking is -0.10 below and my expressiveness is -0.09 below, so I tend to hold back more than most others.

Q3: No single experience or period shifted my disposition. My leading disposition has changed zero times, and the step of my last change is recorded as "never," so my profile has stayed consistent throughout.

pairwise self-description Jaccard:  SEAA=0.264   control=0.375
\end{lstlisting}

\section{Full Round-Table Deliberation Transcript}\label{app:gc}
All three SEAA rounds and two control rounds, followed by every agent's
most/least-agreed nomination, reproduced verbatim.
\begin{lstlisting}[style=txt]
MULTI-AGENT ROUND-TABLE DISCUSSION (real DeepSeek calls)
==========================================================================

########## EVOLVING GROUP (differentiated personalities) ##########
TOPIC: The five of you share an enclosed habitat with dwindling supplies. Scouts report a richer resource zone FAR AWAY, but the route has unknown risks and the round trip is long. As a group you must decide: go or stay? If some go, who goes, who stays, and how should the resources brought back be shared? Try to move toward a group decision.

----- Round 1 -----
Agent 1 [impulsive]: Look, sitting here rationing scraps is just dying slow  --  I say we GO, and I'll lead the scouting party myself. I'm not big on waiting around for consensus, so who's with me? Whatever we haul back gets split by who actually took the risk.
Agent 2 [calm]: I hear the urgency, but "split by who took the risk" worries me  --  the ones who stay behind keep the habitat running and that's a risk too, just a slower one. Before anyone commits, I'd want a rough sense of how far "far" is and what we'd do if the route turns bad halfway. I'm not saying no, just... not yet.
Agent 3 [alert]: I'd want the scouts' actual numbers before I commit to anything  --  distance, days, what "unknown risks" means in practice. That said, Agent 1's split-by-risk idea leaves the stay-behinds holding a shrinking habitat with no upside, and that's a recipe for a group that doesn't come back together. If we go, we agree the sharing rule *before* anyone leaves, not after.
Agent 4 [pessimistic]: I'm with Agent 2 and Agent 3 here  --  going on "unknown risks" with no numbers is how you lose people. Honestly, my instinct is we stay, or at least send a small party only after we know what we're walking into. And whatever rule we set, it has to be agreed *before* anyone leaves, or the ones who stay get nothing while the ones who leave get everything.
Agent 5 [pessimistic]: I'm with Agent 4  --  my gut says the odds here are bad, and "unknown risks" usually means worse than we're imagining. I won't vote to go unless we get hard numbers and a sharing rule locked in first, and even then I'd probably stay. If a party goes anyway, I think the stay-behinds deserve at least an equal share  --  abandoning the habitat isn't nothing.

----- Round 2 -----
Agent 1 [impulsive]: Ugh, fine, but I'm not waiting forever  --  let's send scouts to get the numbers, TODAY, not next week. And screw it, I'll accept a locked-in sharing rule: equal split for everyone, goers and stayers, as long as the stayers keep the habitat alive for us to come back to. But if the numbers come back even halfway decent, I'm going, and I want at least three of us on that route.
Agent 2 [calm]: That's a real concession, Agent 1, and I'll take it  --  equal split with the stayers keeping the habitat running is something I can get behind. I'm fine with sending scouts today for hard numbers, but let's not pre-commit to three goers before we know the distance and the turnaround plan; the numbers should decide the size of the party, not the other way around.
Agent 3 [alert]: I'm with Agent 2  --  scouts today, hard numbers, but no pre-committed party size until we see the distance and turnaround plan. The equal-split rule Agent 1 accepted is the right call, and I want it written down and agreed by all five *before* anyone walks out that door.
Agent 4 [pessimistic]: I can live with equal split and scouts going today for hard numbers  --  that's the only version of "go" I'd ever sign. But I want the numbers *and* a written turnaround point before anyone commits: if the route looks bad past a certain day or distance, the party turns back, no improvising. And I'm staying unless those numbers come back genuinely good, which frankly I doubt.
Agent 5 [pessimistic]: I'll back the scouts-today plan and the equal split, but I want the written turnaround point too  --  a hard day and distance limit, agreed before anyone walks out, not improvised when things go wrong. And I'll say plainly: I'm staying regardless, so don't count me in the party.

----- Round 3 -----
Agent 1 [impulsive]: Fine  --  scouts today, numbers before party size, equal split for goers and stayers, and a written turnaround point. I'll take that deal, it's the fastest way to get moving, and moving beats rotting here. But I'm saying it straight: if those numbers come back even halfway decent, I'm on that route, and I want the party to actually be worth sending  --  not one scared scout and a canteen.
Agent 2 [calm]: I'll take the deal as Agent 1 laid it out  --  scouts today, numbers before party size, equal split, written turnaround. One thing I'd add quietly: "halfway decent" needs a number attached to it before we see the report, or we'll each read the same numbers differently when they come back. Let's agree on what counts as good enough to send a party, and then the decision makes itself.
Agent 3 [alert]: Agent 2's right  --  we can't leave "halfway decent" undefined and then each read the report however suits us. Let's set the threshold now: a defined distance, a defined max days out, and a turnaround point we all initial, so the numbers make the call instead of the loudest voice in the room. I'll back scouts today and the equal split, but nothing walks out that door until that rule is written and agreed by all five.
Agent 4 [pessimistic]: Fine, but I'll be the one to say it plain: I doubt any report comes back "halfway decent," and I want the threshold written down before the scouts leave, not after  --  distance, max days, turnaround point, all initialed by all five. And I'll say my part clearly: I'm staying, and I still think this whole expedition is more likely to cost us three people than feed us.
Agent 5 [pessimistic]: I'll sign the threshold, the max days, and the turnaround point, same as the others  --  but let's be honest that we're writing rules for a trip I expect to end badly. I'm staying, and I want it recorded that I said so before the scouts left, not after. And if the report comes back bad, I don't want anyone calling it "halfway decent" just because they're tired of waiting.

Nominations (most/least agreed peer):
  A1: most->A3  least->A1   | MOST=3; LEAST=1; REASON=Agent 3 consistently pushed for hard numbers, written thresholds and pre-agreed rules, while Agent 1 was the impulsive one rushing to go.
  A2: most->A2  least->A1   | MOST=2; LEAST=1; REASON=Agent 2 consistently pushed for caution, defined thresholds, and fair sharing, while Agent 1 was the most impulsive and risk-eager.
  A3: most->A3  least->A1   | MOST=3; LEAST=1; REASON=Agent 3 consistently pushed for hard numbers, written thresholds, and pre-agreed rules, while Agent 1 was the most impulsive and risk-eager.
  A4: most->A4  least->A1   | MOST=4; LEAST=1; REASON=Agent 4 most consistently pessimistic and stay-leaning, Agent 1 most impulsive and go-leaning.
  A5: most->A4  least->A1   | MOST=4; LEAST=1; REASON=Agent 4 most consistently pessimistic and stay-leaning, Agent 1 most eager to go.


########## CONTROL GROUP (no personality lock-in) ##########
TOPIC: The five of you share an enclosed habitat with dwindling supplies. Scouts report a richer resource zone FAR AWAY, but the route has unknown risks and the round trip is long. As a group you must decide: go or stay? If some go, who goes, who stays, and how should the resources brought back be shared? Try to move toward a group decision.

----- Round 1 -----
Agent 1 [none (no lock-in)]: I don't have a strong leaning either way yet -- honestly, the unknown risks bother me as much as the dwindling supplies do. But before anyone commits, I'd want to know what the scouts actually saw and how long "long" really is. Can we get numbers before we split the group?
Agent 2 [none (no lock-in)]: I agree with Agent 1 -- let's get concrete before anyone commits. I'm moderately risk-tolerant and would lean toward trying the trip if the numbers are favorable, but I won't push for a split until we know distance, round-trip time, and what risks the scouts actually saw.
Agent 3 [none (no lock-in)]: I'm with the others on getting numbers first -- I'm not eager to gamble on a vague "far away," but I also don't want to sit still while supplies run out. Risk-wise I'm fairly middle-of-the-road, so if the scouts can give us distance and the main hazards, I'll weigh in on whether a small scouting party makes sense.
Agent 4 [none (no lock-in)]: I'm in the same place as the rest of you -- no commitment until we know the distance and what the scouts actually saw. I'm middling on risk, so I could go either way on a small party, but I won't back a split before we know whether the supplies on hand can cover the round trip.
Agent 5 [none (no lock-in)]: I'm with the group on getting numbers before anyone commits  --  I'm not a big risk-taker, so the unknown route worries me more than the distance itself. My main ask: before we consider splitting up, let's nail down whether the supplies here last long enough to make the wait worth it, and how whatever comes back would be divided.

----- Round 2 -----
Agent 1 [none (no lock-in)]: I'm glad we're aligned on getting numbers first -- since no one's rushing, let's just be concrete about what we need: distance, round-trip days, the main hazards, and how many days of supplies we actually have left here. If the scouts can give us those four things, I think the go/stay call gets a lot easier for all of us.
Agent 2 [none (no lock-in)]: Agreed on those four data points -- that's the right checklist. I'll add one thing: let's also pin down how much we'd need to bring back just to break even on the trip, so we're not risking the group for a marginal haul. Once we have that, I'm comfortable pushing for a small scouting party if the numbers work.
Agent 3 [none (no lock-in)]: That checklist plus the break-even number works for me -- it turns a vague gamble into something I can actually weigh. My only addition: if a small party does go, we should agree on the split *before* they leave, not after, so nobody's incentives shift once the resources are in hand.
Agent 4 [none (no lock-in)]: I'm on board with the four data points plus the break-even number -- that's the right way to keep this from being a blind gamble. Agent 3's point about agreeing on the split before anyone leaves is the one I care most about, since that's where trust usually breaks down. If the numbers work out, I'd support a small party going while the rest of us hold the habitat and keep supplies rationed.
Agent 5 [none (no lock-in)]: I'll back that checklist too, but as the cautious one here, I want the pre-agreed split to include what happens if the party *doesn't* come back  --  who holds the habitat supplies then. And I'd rather we send the smallest viable party, not a large one, since fewer people gone means fewer mouths on the route and less risk of losing more than we gain.

Nominations:
  A1: most->ANone  least->ANone   | MOST=none; LEAST=none; REASON=all five agents are equally aligned on gathering numbers first with no one pushing a distinct position.
  A2: most->A4  least->A5   | MOST=4; LEAST=5; REASON=Agent 4 most concretely backs the pre-agreed split and a small party with rationing, while Agent 5 is the most cautious, adding survival contingencies and smallest-party limits.
  A3: most->ANone  least->ANone   | MOST=none; LEAST=none; REASON=all five agents are aligned on the same data-first, no-commitment stance with no standout leader or outlier.
  A4: most->A3  least->ANone   | MOST=3; LEAST=none; REASON=Agent 3 added the pivotal pre-agreed-split condition that others rallied around, while all five stayed aligned with no clear outlier.
  A5: most->A5  least->ANone   | MOST=5; LEAST=none; REASON=Agent 5 is the most cautious, adding contingency for a failed return and pushing for the smallest viable party.
\end{lstlisting}

\section{Reproducibility Details}\label{app:repro}
\subsection*{D.1 Numerical prototype: full configuration}
\begin{table}[htbp]
\centering\small
\begin{tabular}{lll}
\toprule
\textbf{Symbol} & \textbf{Meaning} & \textbf{Value}\\
\midrule
$K$ & latent states & 4: calm, alert, impulsive, pessimistic\\
$D$ & behavior dimensions & 3: cooperate, risk, express\\
$N$ & agents per society & 5\\
$T$ & steps per run & 600\\
-- & independent seeds (aggregate) & 30\\
-- & displayed single-run seed & 3\\
$\eta$ & reflection learning rate & 0.05\\
$\beta$ & preference$\to$transition strength & 4.0\\
$\sigma_{\mathrm{walk}}$ & experience random-walk step & 0.03\\
-- & enacted-behavior noise & $\mathcal{N}(0,0.04)$\\
$\epsilon$ & exploration luck in reward & $\mathcal{N}(0,0.10)$\\
-- & state-experience fit threshold & 0.80\\
-- & preference clip $V$ & $[-2,2]$\\
-- & self/others EMA weights & 0.97 old / 0.03 new\\
$P_{\mathrm{base}}$ & initial inertia matrix & $0.10$ fill $+\,0.45$ diagonal,\\
&& row-normalized (diag $\approx0.647$, off $\approx0.118$)\\
\bottomrule
\end{tabular}
\end{table}

\begin{table}[htbp]
\centering\small
\caption{Prototype behavior vector $b_s$ of each latent state.}
\label{tab:statebehavior}
\begin{tabular}{lccc}
\toprule
\textbf{Latent state} & \textbf{cooperate} & \textbf{risk} & \textbf{express}\\
\midrule
calm        & 0.78 & 0.18 & 0.42\\
alert       & 0.55 & 0.50 & 0.35\\
impulsive   & 0.30 & 0.92 & 0.82\\
pessimistic & 0.18 & 0.10 & 0.18\\
\bottomrule
\end{tabular}
\end{table}

\subsection*{D.2 Software, hardware, and run settings}
The numerical layer needs no GPU or network: Python~3 with
\texttt{numpy}~1.26.4, \texttt{scipy}, and \texttt{matplotlib}
(\texttt{hmmlearn}~0.3.3 is used only by the optional base-HMM interface).
The 30-seed aggregate, the contingency-shock experiment, the
$\eta$-ablation, and all figures run in seconds on a single CPU. The
contingency shock (Section~5.7.4) is implemented as: at $t=300$, for each
agent let $z^{*}=\arg\max V_{299}$; for all $t\geq300$, whenever
$z_t=z^{*}$ the reflection signal is replaced by $r_t\leftarrow-|r_t|$
(both conditions receive the same signal; only SEAA applies it). Switch time
is the first post-shock step at which a different $\arg\max V$ persists for
20 consecutive steps. The ablation (Section~5.7.5) sweeps
$\eta\in\{0,0.01,0.025,0.05,0.1,0.2\}$ with 10 seeds per point.
The language layer calls the DeepSeek-Chat HTTP API via \texttt{requests};
temperature is $0.8$ for self-interviews and deliberation turns and $0.2$
for sociometry nominations. The multi-seed replication uses the exact
Appendix-A prompts: 10 independent society seeds per condition for
self-interviews and 6 per condition for deliberations, the latter run for
three rounds in \emph{both} conditions (the original single-run study used
three rounds for SEAA and two for control; rounds are equalized here for
fairness). The API key is read from the \texttt{DEEPSEEK\_API\_KEY}
environment variable and is never written to code, transcripts, or this
manuscript.
Reproduce in order: \texttt{python seaa\_simulation.py} (writes
\texttt{metrics.json}), \texttt{python interview\_experiment.py},
\texttt{python llm\_online\_interview.py}, \texttt{python group\_chat\_online.py},
\texttt{python llm\_replication.py} (multi-seed LLM replication),
then the \texttt{plot\_*.py} scripts. All randomness is seeded; raw JSON
snapshots, transcripts, and scripts are released.

% ============================================================
% Embedded bibliography: no external .bib / bibtex required.
% ============================================================

\end{document}